# KonvLiNA: Integrating Kolmogorov-Arnold Network with Linear Nyström Attention for feature fusion in Crop Field Detection

*Haruna Yunusa [1], Prof. Qin Shiyin [1], Adamu Lawan [1], Abdulrahman Hamman Adama Chukkol [2]

[1] Beihang University, Beijing, China
[2] Beijing Institute of Technology, Beijing, China

## ABSTRACT

Crop field detection is a critical component of precision agriculture, essential for optimizing resource allocation and enhancing agricultural productivity. This study introduces KonvLiNA, a novel framework that integrates Convolutional Kolmogorov-Arnold Networks (cKAN) with Nyström attention mechanisms for effective crop field detection. Leveraging KAN adaptive activation functions and the efficiency of Nyström attention in handling large-scale data, KonvLiNA significantly enhances feature extraction, enabling the model to capture intricate patterns in complex agricultural environments. Experimental results on rice crop dataset demonstrate KonvLiNA superiority over state-of-the-art methods, achieving a 0.415 AP and 0.459 AR with the Swin-L backbone, outperforming traditional YOLOv8 by significant margins. Additionally, evaluation on the COCO dataset showcases competitive performance across small, medium, and large objects, highlighting KonvLiNA efficacy in diverse agricultural settings. This work highlights the potential of hybrid KAN and attention mechanisms for advancing precision agriculture through improved crop field detection and management.

***Keywords.*** Crop Field Detection, Feature Fusion, Kolmogorov-Arnold Networks (KAN), Nyström Attention Mechanism, Object Detection

## 1. INTRODUCTION

Agricultural field disease detection is crucial due to the significant threat these diseases pose to global food security, requiring timely and accurate identification for effective management and mitigation [1]. Advances in imaging technologies and artificial intelligence have enabled the development of automated detection systems, offering more efficient and precise solutions, unlike traditional methods which rely on expert knowledge and manual inspection, which are time-consuming and error-prone [2].

Notable advancements in deep learning have significantly enhanced the capabilities of computer vision systems in agriculture. Convolutional neural networks (CNN) and attention mechanisms have shown great promise in extracting and processing visual information for various applications [3, 4]. However, not enough exploration has been done in the agricultural domain to specifically design modules that handle the challenge of scale variations and enhance the capturing of intricate details prevalent in complex open fields, such as farms. Conventional CNN pyramid feature fusion struggles with capturing multi-scale objects due to its hierarchical structure that leads to high inductive bias, while standard attention mechanisms, despite having low inductive bias, can be computationally intensive [6].

Recently, a novel architecture based on the Kolmogorov-Arnold representation theorem, namely, Kolmogorov-Arnold Network (KAN) [8], has emerged as promising alternative to Multi-Layer Perceptron (MLP). Unlike MLP, which have fixed activation functions on nodes, KAN utilize learnable activation functions on edges, replacing linear weights with univariate functions parametrized as splines. This structural change allows KANs to outperform MLP in terms of both accuracy and interpretability, particularly in tasks requiring fine-grained function fitting. Extending this concept to convolutional operations, KAN Convolutions operate similarly to traditional convolutions but with a critical difference, instead of applying a dot product between the kernel and the corresponding pixels in the image, KAN Convolutions apply a learnable nonlinear activation function to each element before summing them up. The kernel of a KAN Convolution is equivalent to a KAN linear layer with four inputs and one output neuron. For each input $i$, a learnable function $\varphi_i$ is applied, and the resulting pixel of that convolution step is the sum of $\varphi_i(x_i)$. This process enhances the network capacity to model complex function and capture intricate patterns in the data [8].

To further enhance feature extraction and integration in complex agricultural landscapes, we introduce Linear Nyström Attention mechanism. The traditional self-attention mechanism, while powerful, can be computationally costly when applied to large-scale images, due to its quadratic complexity in terms of input size [5]. The Nyström method, a well-established technique in numerical linear algebra, approximates large-scale kernel matrices with a smaller subset of columns, significantly reducing computational overhead while maintaining a high level of accuracy [7]. By integrating Linear Nyström Attention with KAN Convolutions, KonvLiNA, achieves a synergistic effect, combining the fine-grained, non-linear function modeling capabilities of KAN with the efficient and scalable attention mechanism of the Nyström method. This integration not only addresses the computational challenges posed by large-scale agricultural images but also enhances the model's ability to capture both local and global features essential for accurate crop field detection.

To this end, we propose KonvLiNA, a pyramid feature fusion approach for enhancing object detection in agricultural fields. The approach integrates two novel modules: Convolutional KAN Spatial Pyramid Pooling (cKSPP) and Enhanced Nyström Attention Upsample (ENAU). The former effectively leverages the expressive power of KAN to capture the complex, intricate details prevalent in large open fields like farms, while the latter utilizes the Linear Nyström Attention mechanism to address information loss, redundancy, and degradation of feature maps during the up-sampling process. Additionally, ENAU mitigates artifacts present during training and inference with register tokens. The proposed hybrid KonvLiNA network not only enhances the ability to capture complex, intricate details but also improves the capturing of long-range dependencies prevalent in large, open farm fields, where objects of interest may be tiny.

The contributions of this study are summarized below:

- Designed a novel cKSPP module capable of efficiently capturing the complex, intricate details prevalent in agricultural fields.
- Designed a novel ENAU module, which mitigates information loss and artifacts during training and inference.
- Extensive experimental results demonstrate promising performance compared to some state-of-the-art pyramid feature fusion approaches.

## 2. RELATED WORK

Existing advanced pyramid feature fusion modules for object detection utilize either conventional CNNs, attention mechanisms, or a hybrid of both approaches to address the challenge of capturing multi-scale variations in object detection tasks.

**CNN Based.** Asymptotic Feature Pyramid Network (AFPN) enhances object detection by enabling direct interaction between non-adjacent pyramid levels, initially fusing low-level features and progressively integrating higher-level features to minimize semantic gaps. Its adaptive spatial fusion effectively handles inconsistencies, resulting in improved average precision and computational efficiency on the MS COCO 2017 dataset compared to traditional FPNs [6]. The Parallel Residual Bi-Fusion Feature Pyramid Network (PRB-FPN) introduces a bi-directional feature fusion approach, utilizing a Bottom-Up Fusion Module and a Concatenation and Re-Organization module with residual design for high-quality single-shot object detection. This method achieves state-of-the-art results on UAVDT17 and MS COCO datasets [9]. EfficientDet improves object detection through the weighted bi-directional feature pyramid network (BiFPN) for efficient multi-scale feature fusion and a compound scaling method. It improves accuracy with fewer parameters and FLOPs, with EfficientDet-D7 reaching 55.1 AP on the COCO test-dev [26]. RevBiFPN, a reversible bidirectional feature pyramid network, reduces memory requirements during training by recomputing hidden activations. It shows competitive performance with EfficientNet while using significantly less memory, making it suitable for large-scale models [16].

**Attention and Hybrid Based.** The Pyramid Attention Object Detection Network utilizes a multi-scale feature fusion pyramid attention module, enhancing the detection of small and partially occluded objects by integrating global average pooling results from multiple scales. This approach significantly improves detection accuracy on PASCAL VOC and MS COCO datasets [11]. ReAFFPN enhances rotation-equivariant feature fusion in aerial object detection through Rotation-equivariant Channel Attention, improving classification accuracy and feature fusion consistency. This method significantly boosts the accuracy of Rotation-equivariant Convolutional Networks (ReCNNs) [12]. RHF-Net addresses small object detection on embedded devices by introducing a bidirectional fusion module and a recursive concatenation and reshaping module. It enhances detection accuracy and efficiency on COCO and UAVDT datasets [10]. The refined marine object detector, with attention-based spatial pyramid pooling networks and a bidirectional feature fusion strategy, enhances feature representation and detection accuracy in underwater environments, achieving high mAP on underwater image and URPC datasets [13]. The Attentional Feature Fusion module leverages attention mechanisms for superior feature integration, enhancing performance across various CNN architectures and achieving state-of-the-art results on CIFAR-100 and ImageNet [14]. The Feature Pyramid Network with Multi-Scale Prediction Fusion for real-time semantic segmentation uses a dual prediction module and attention mechanism to enhance segmentation accuracy and speed, achieving notable performance on Cityscapes and Mapillary Vistas datasets [15].

**Discussion.** While these advanced methods significantly improve object detection accuracy through enhanced multi-scale feature fusion and contextual information preservation, they primarily focus on general object detection tasks. There is a lack of exploration in specific applications like agriculture, where unique challenges such as varying scales and complex backgrounds are prevalent. To address these challenges, we will explore integrating KAN to leverage its dynamic learnable activation function to capture richer semantics prevalent in crop fields with Linear Nyström Attention mechanism to efficiently capture long-range dependencies and preserve contextual information due to the object of interests spanning the whole image. This integration could lead to a more robust and accurate detection systems custom in agricultural applications, contributing significantly to the field of precision agriculture.

## 3. METHOD AND TOOLS

### 3.1 KonvLiNA Overview

This section provides a general overview of KonvLiNA, which is based on a multi-scale fusion of Convolutional Kolmogorov-Arnold Network and up-sampling Linear Nyström Attention mechanism with Registers. High and low-level features extracted from the backbone networks are fed into KonvLiNA to extract multi-scale information from images leveraging the expressive power of KAN using the cKSPP module. Then, eNAU to mitigate information degradation of spatial details during the fusion. Finally, fusing the top-down and lateral connections and the output is sent to the detection head. The remainder of this section presents the components used in the KonvLina framework. Refer to Figure 1 for illustrations of all interactions between each component.

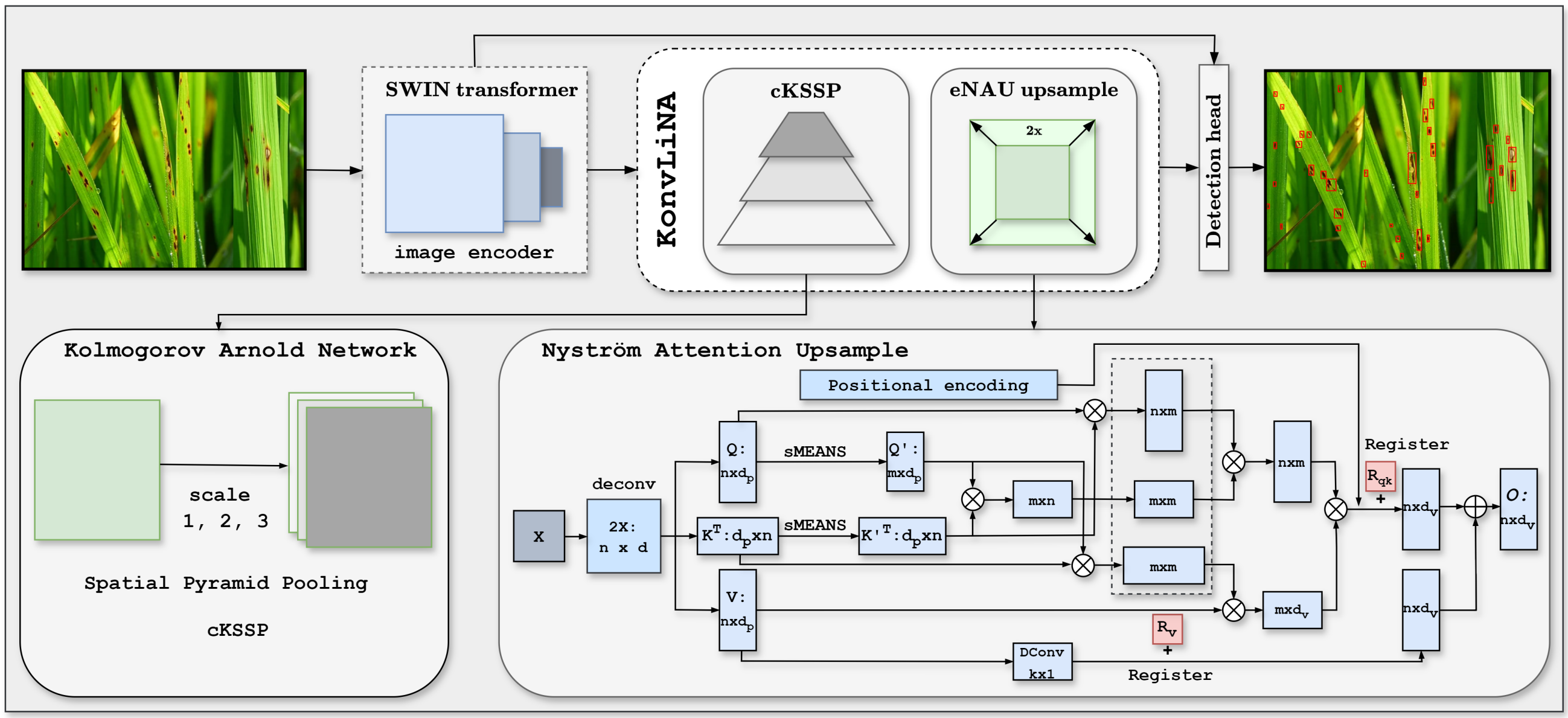

Figure 1. KonvLina Architectural design

### 3.2 Image Encoder

For precise crop field detection, KonvLiNA uses the Swin Transformer [16], a hierarchical image encoder that efficiently captures long-range dependencies and retains both low-level and high-level features by processing images in shifted local windows. The input $X \in \mathbb{R}^{H \times W \times C}$ is partitioned into non-overlapping windows of size $M \times M$, where self-attention is computed within each window. These windows are then shifted by a fixed offset $s$ to capture cross-window interactions, and features are aggregated and down-sampled hierarchically. Multi-scale feature maps $F_1, F_2, F_3$ are extracted from different layers, making the Swin Transformer a powerful image encoder.

### 3.3 Convolutional Kolmogorov Arnold Network

Kolmogorov-Arnold representation theorem states that any multivariate continuous function can be expressed as finite sum of compositions of univariate function [8], eqn. 1.

$$f(x_1, x_2 \dots, x_n) = \sum_{q=1}^{2n+1} \Phi_q \left( \sum_{p=1}^{n} \emptyset_{q,p}(x_p) \right) \tag{1}$$

where $\Phi_q$ and $\emptyset_{q,p}$ are continuous functions mapping each input variable, $x_p$ and $\Phi_q$ respectively. This allows KAN to represent complex relationships in high-dimensional data by combining univariate functions. It utilizes this theorem by replacing traditional linear weights with spline-parametrized univariate functions, using adaptive, learnable activation functions on edges between nodes with b-spline curves that adjust during training, which allows it to capture complex nonlinear relationships more effectively than traditional MLPs with non-learnable activation functions, eqn. 2 is a deeper KAN architecture.

$$KAN(x) = (\Phi_{L-1} \circ \Phi_{L-2} \circ \dots \circ \Phi_0)(x) \tag{2}$$

where each $\Phi_L$ denotes a KAN layer. The number of layers enables the detection of more complex patterns, with each layer $l$ applying a sequence of learnable functions $\emptyset_{q,p}$, making the network flexible and robust.

**Convolutional KAN.** [18] is inspired by the standard CNN architectures, requiring fewer parameters due to the use of b-splines. These splines offer smoother activation function representations compared to ReLU. In KAN convolutions, the implementation diverges from traditional CNN convolutions primarily in the nature of the kernel employed. While CNNs utilize weight-based kernels, Convolutional KANs operates kernels where each element, $\emptyset$, is a learnable non-linear function utilizing b-splines, eqn. 3

$$\emptyset = w_1 \cdot spline(x) + w_2 \cdot silu(x) \tag{3}$$

In KAN Convolutions, the kernel traverses the image, applying the activation function $\emptyset_{ij}$ to each pixel $a_{kl}$ and computing the output pixel as the sum of $\emptyset_{ij}(a_{i+k,j+l})$. Formally, if $K$ represents a KAN kernel in $\mathbb{R}^{N \times M}$, then the image is represented as a matrix.

$$image = \begin{bmatrix} a_{11} & a_{12} & \dots & a_{1p} \\ a_{21} & a_{22} & \dots & a_{2p} \\ \vdots & \vdots & \ddots & \vdots \\ a_{m1} & a_{m2} & \dots & a_{mp} \end{bmatrix}$$

The KAN convolution is defined as, $(image * K) = \sum_{k=1}^{N} \sum_{l-1}^{M} \emptyset_{ij}(a_{i+k,j+l})$. This approach effectively integrates the flexibility of KANs with the spatial processing capabilities of convolution operations, enhancing the model's ability to capture complex spatial dependencies in data.

### 3.4 cKSPP (convolutional KAN Spatial Pyramid Pooling)

We propose cKSPP by leveraging the expressive power of convolutional KAN using its dynamic learnable activation function to capture richer semantics prevalent in crop open fields images. Our approach defines a multi-scale pooling to the outputs of the image encoder $F_1, F_2, F_3$. Then we apply various pooling scales in order to encode multiple scale features. Adaptive average pooling changes the size of the input feature map to a fixed output size (s, s), where s represents the scale, for an input $X \in \mathbb{R}^{H \times W \times C_{in}}$, eqn. 4 formulates the spatial pooling operation.

$$X_{pool}^{(s)} = AvgPool2D(X) \in \mathbb{R}^{s \times s \times C_{in}} \tag{4}$$

Where $X_{pool}^{(s)}$ is the pooled feature map at scale s. Then we applied KAN layer to $X_{pool}^{(s)}$ capture complex pattern prevalent in open fields farms. KAN has a learnable activation function which gives it better expressiveness using it splines curves which adjust during training. The operation is formulated in eqn. 5.

$$X_{KAN}^{(s)} = KAN_{1 \times 1}(X_{pool}^{(s)}) \in \mathbb{R}^{s \times s \times C_{cout}} \tag{5}$$

cKSPP captures richer semantics by combining multi-scale pooling with KAN dynamic learnable activation function. This approach enhances the expressiveness of the model, enabling it to detect complex patterns across various scales, particularly in challenging environments like open-field farms. By adapting feature maps to different scales and refining them through KAN, cKSPP improves the model's ability to handle diverse and intricate visual data.

### 3.5 eNAU (enhanced Nyström Attention Up-sample)

In this section, we propose a novel up-sampling method using a Linear Nyström Attention to mitigate information loss, degradation of feature maps, and reduce artifacts in KonvLiNA. Given a normalized feature map $X \in \mathbb{R}^{L\times C}$ (where $L$ represents the length of the sequence) generated by the cKSPP module, we first apply a $1 \times 1$ convolution for dimensionality reduction, ensuring the feature map channel dimension is appropriately scaled for subsequent processing. Next, we generate the query $Q$, key $K$, and value $V$ matrices from input $X$. We then apply the Nyström method to approximate the self-attention mechanism, which enables more efficient computation. The Nyström approximation utilizes a subset of the sequence data to estimate the full attention matrix, reducing the computational complexity while maintaining the benefits of attention. Then we compute low-rank approximations of the attention matrix by projecting $Q$ and $K$ onto a lower-dimensional space. This approximation allows us to efficiently compute attention scores and apply them to $V$, resulting in attention outputs with reduced computational complexity to $O(N \log N)$. Refer to eqn. 6 for Nyström Approximation.

To preserve sequence order, we add relative positional encoding. The learnable positional encodings are initialized with random values $p_{rel} \in \mathbb{R}^{h\times D_h\times L}$, where $h$ is the number of heads and $D_h$ is the dimension of each head. This encoding matrix $P = p_{rel}$ is used to integrated positional information into the attention mechanism. We utilize learnable deconvolutional layers for up-sampling to better preserve fine details and adaptively enhance quality, unlike nearest neighbor interpolation, which is non-trainable and can lead to redundant up-sample feature maps. This effectively up-samples the sequence length, resulting in the projection of the attention outputs with higher resolution.

Lastly, to address the problem of artifacts introduced by attention mechanisms and deconvolutions, we introduce registers [19]. The registers are added to $Q$ and $K$ as $(Q, K)_R = (QK) + R$ and to $V$ as $V_R = V + R$, where $R$ is trainable token with the same size as $QKV$ vectors. These registers are discarded after computing the Linear Nyström Attention during training and inference.

### 3.6 Linear Nyström Approximation

The Nyström method approximates the attention mechanism by selecting a subset of $m$ landmarks from $K$. Let $K_m \in \mathbb{R}^{m\times d_k}$ be the subset of landmarks, and $Q_m$ be the corresponding queries. The approx. is computed as in eqn. 6.

$$\text{Nyström}\ (Q, K, V) \approx \text{softmax}\left(\frac{QK_m^T}{\sqrt{d_k}}\right)\left(\text{softmax}\left(\frac{K_m K_m^T}{\sqrt{d_k}}\right)\right)^{-1} \text{softmax}\left(\frac{K_m K^T}{\sqrt{d_k}}\right)V \tag{6}$$

This approximation reduces the complexity of self-attention from $O(N^2)$ to $O(N \log N)$, making it more efficient for large sequences while maintaining similar performance.

### 3.7 Feature Fusion

KonvLiNA module combines eNAU and cKSPP through a pair-wise addition fusion operation. eNAU utilizes a Linear Nyström Attention mechanism to reduce information loss and mitigate homogeneous semantic information, while cKSPP applies custom spatial pyramid pooling for multi-scale feature extraction, capturing intricate details of small objects in images. The cKAN layer further enhances the module by capturing complex patterns typical in open-field farms, leveraging KAN expressive power through its learnable activation function. The combined output is then fed into a YOLOv8 detection head, which operates across three pyramid levels to effectively detect objects in open farm environments. This approach ensures robust detection by preserving spatial details and capturing diverse features, both crucial for agricultural applications.

## 4 EXPERIMENTAL RESULTS AND EVALUATION

In this section, we report the experimental results and evaluate KonvLiNA both quantitatively and qualitatively to assess its effectiveness in enhancing crop fields object detection using two main datasets.

**Dataset and metric.** The Mendeley Rice Leaf Disease dataset [20] was used, which consists of 5,932 images of four disease types: Brown Spot, Bacterial Blight, Blast, and Tungro. This dataset was used to validate the effectiveness of our method and compare it with existing methods. Additionally, to evaluate the robustness of our method, we utilized the COCO dataset, which consists of 330,000 images of 80 object categories. Average Precision (AP) and Mean Average Precision (mAP) were used to evaluate the model's performance.

**Experimental setup.** The input images were set to 640 × 640 pixels pixels for both training and evaluation over 90 epochs using the YOLOv8 network settings. The bias values for the classification and localization layers in the detection head were set to 0.01 and 0.1, respectively. A Gaussian weight with σ = 0.01 was used in all layers, including the proposed feature selection network. We used Adam optimizer with an initial learning rate of 0.001, weight decay of 0.0005, and momentum of 0.95. Data augmentation included 0.5 horizontal/vertical flip, mosaic of 0.9, and scale of 0.5. Due to the small training dataset, we fine-tuned with the MS-COCO pre-trained weights.

### 4.1 Quantitative Comparison

This section evaluates the performance of YOLOv8 with KonvLiNA and with other feature pyramid fusion models with Mendeley RLD crop dataset. In addition, we evaluated it with COCO dataset for robustness.

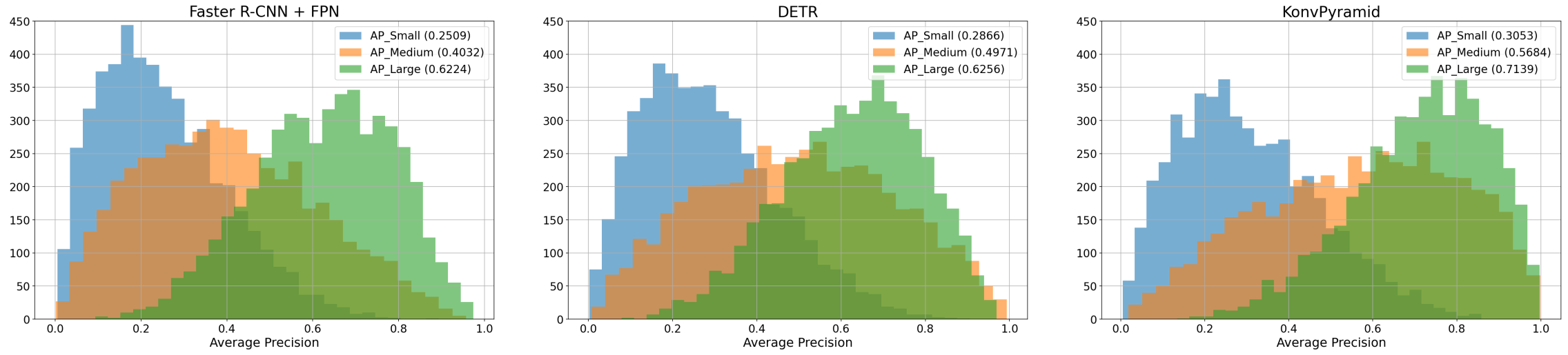

Faster R-CNN + FPN DETR KonvLiNA

Figure 2 shows the comparative analysis of AP values across small, medium, and large objects which reveals notable performance enhancements in DETR and KonvLiNA architectures relative to Faster R-CNN + FPN on the RLD dataset. DETR shows improvements of approximately 14.2% for small objects, 23.3% for medium objects, and a marginal 0.5% for large objects compared to Faster R-CNN. In contrast, KonvLiNA exhibits more substantial gains with approximately 21.7% improvement for small objects, 41.0% for medium objects, and 14.7% for large objects. These results highlight the effectiveness of KonvLiNA in handling scale variations across all object sizes compared to Faster R-CNN + FPN and DETR.

**Rice Crop Disease Dataset.** Utilizing the Swin-L backbone, KonvLiNA outperforms all other methods with improvements of 3.50% in AP and 4.37% in AR over YOLOv8, 0.75% in AP and 2.94% in AR over DETR, 0.73% in AP and 3.79% in AR over HTC++, 2.30% in AP and 4.20% in AR over ViT Det, and 0.08% in AP and 3.80% in AR over DINO. Overall, these results demonstrate significant performance enhancements in detecting objects with various variations prevalent in Rice Crop fields, where objects of interest are typically small due to the size of rice leaves. This highlights the potential benefits of exploring hybrid KAN and attention mechanisms in such scenarios.

Table 1 AP comparison on Rice Crop disease detection

| **Object Detector** | **Backbone** | **AP** | **AR** |
|---|---|---|---|
| Faster R-CNN+FPN | ResNet-101 | 0.372 | 0.402 |
| YOLOv8 | Darknet-53 | 0.380 | 0.415 |
| DETR | Swin | 0.408 | 0.430 |
| HTC++ | Swin V2-G | 0.410 | 0.422 |
| ViTDet | BEiT-3 | 0.392 | 0.417 |
| DINO | Swin-L | 0.407 | 0.421 |
| KonvLiNA (ours) | Swin-L | 0.415 | 0.459 |

**COCO dataset.** On the COCO dataset, KonvLiNA achieves competitive AP across small, medium, and large objects compared to various Faster R-CNN configurations with different FPN.

Table 2 AP comparison on COCO dataset

| **Method** | $AP$ | $AP_{50}$ | $AP_{75}$ | $AP_S$ | $AP_M$ | $AP_L$ |
|---|---|---|---|---|---|---|
| Faster R-CNN + FPN [21] | 39.7 | 61.4 | 43.3 | 22.3 | 42.9 | 50.4 |
| Faster R-CNN + PAFPN [22] | 38.1 | 58.1 | 41.3 | 19.1 | 42.5 | 54.0 |
| Faster R-CNN + Graph FPN [23] | 39.1 | 58.3 | 39.4 | 22.4 | 38.9 | 56.7 |
| Faster R-CNN + LFPN [24] | 38.7 | 60.4 | 41.9 | 23.5 | 42.5 | 49.0 |
| Faster R-CNN + CARAFE [25] | 38.6 | 59.9 | 42.2 | 23.3 | 42.2 | 49.7 |
| Faster R-CNN + AFPN [6] | 39.0 | 57.6 | 42.1 | 19.4 | 43.0 | 55.0 |
| KonvLiNA (ours) | 39.8 | 59.4 | 43.2 | 22.9 | 44.1 | 56.3 |

### 4.1 Qualitative Comparison

We also provide a qualitative evaluation in Fig. 3, demonstrating that the KonvLiNA module significantly enhances the rice crop detection across diverse object sizes, particularly smaller objects, compared to Faster R-CNN and DETR. e.g., in the upper-left of Fig. 3a, KonvLiNA effectively identifies small disease spots on rice leaves that Faster R-CNN misses. Also, KonvLiNA outperforms Faster R-CNN and DETR in detecting several very small objects in Fig. b. These findings align with our quantitative results, emphasizing KonvLiNA robust performance on small disease spots and its improved accuracy under varying occlusion levels. However, despite these improvements, our model still misses some extremely challenging disease objects, as shown in KonvLiNA Fig. c. right side.

Overall, these qualitative findings highlight the effectiveness of our approach and the significant improvement achieved through detecting small or extremely varied objects in open crop field environments i.e. rice crops.

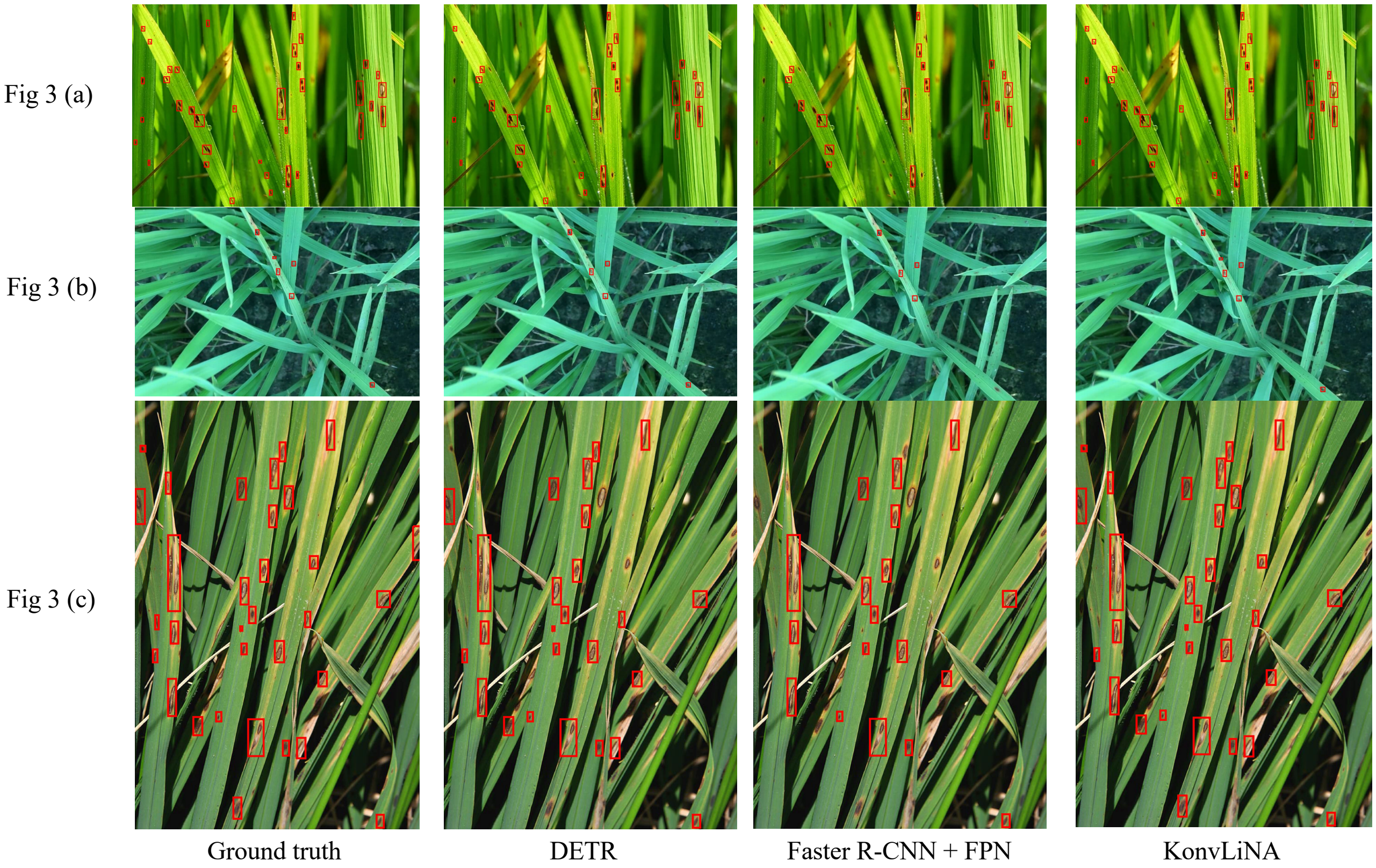

**Ablation.** On the RLD dataset demonstrates that integrating eNAU into the baseline KonvLiNA configuration increases AP by 1.7% (from 0.359 to 0.365) and AR by 3.5% (from 0.371 to 0.384), while including cKSPP alongside eNAU further enhances AP by 15.6% (from 0.359 to 0.415) and AR by 23.7% (from 0.371 to 0.459).

Table 3 Ablation on Rice Crop dataset

| **Components** | **AP** | **AR** |
|---|---|---|
| KonvLiNA + Nearest neighbor | 0.359 | 0.371 |
| KonvLiNA + eNAU | 0.365 | 0.384 |
| KonvLiNA + eNAU + cKSPP | 0.415 | 0.459 |

## 5. CONCLUSION

In this study, we proposed a novel feature pyramid fusion network, KonvLiNA, aimed at improving crop field detection by effectively addressing the challenge of scale variation. Our hybrid module combines the cKAN and Nyström attention mechanisms. Additionally, we introduced eNAU and cKSPP, which reduce information loss during feature fusion and enhance the model's ability to capture intricate patterns through the expressiveness of the KAN activation function. These novelties collectively improve the model's performance in detecting crop fields, demonstrating the effectiveness of our approach.

## AUTHORS' BACKGROUND

| Your Name | Title* | Research Field | Personal website |
|---|---|---|---|
| Haruna Yunusa | PhD | Computer Vision, Deep Learning, Machine Learning | https://yunusa2k2.github.io/portfolio/ |
| Qin Shiyin | Full Professor | pattern recognition and machine learning, image processing and computer vision, and artificial intelligence and knowledge engineering | https://ieeexplore.ieee.org/author/37271751500 |
| Adamu Lawan | PhD | Natural Language Processing, Deep Learning | |
| Abdulrahman Hamman Adama Chukkol | PhD | Computer Vision, Security | |

*This form helps us to understand your paper better, the form itself will not be published.

*Title can be chosen from: master student, Phd candidate, assistant professor, lecture, senior lecture, associate professor, full professor